%% file: acl_latex.tex
\documentclass[11pt]{article}

\usepackage[preprint]{acl}

\usepackage{times}
\usepackage{latexsym}
\usepackage[T1]{fontenc}
\usepackage[utf8]{inputenc}
\usepackage{microtype}
\usepackage{inconsolata}
\usepackage{xurl}
\usepackage{graphicx}

\usepackage{amsmath,amssymb}
\usepackage{algorithm}
\usepackage{algpseudocode}
\usepackage{booktabs}
\usepackage{multirow}
\usepackage{xcolor}
\usepackage{xspace}
\usepackage{pifont}
\newcommand{\cmark}{\textcolor{green!45!black}{\ding{51}}}
\newcommand{\xmark}{\textcolor{red!65!black}{\ding{55}}}
\usepackage{makecell}
\usepackage{array}

\usepackage{tikz}
\usetikzlibrary{arrows.meta,positioning,fit,calc}
\usepackage{fvextra}
\usepackage{enumitem}

\newcommand{\method}{\textsc{GrepMem}\xspace}       
\newcommand{\grep}{\texttt{grep\_history}\xspace}
\newcommand{\bcp}{BrowseComp-Plus\xspace}
\newcommand{\bcpm}{BCP-Multi\xspace}
\newcommand{\cb}{ContextBench\xspace}
\newcommand{\ama}{AMA-Bench\xspace}
\newcommand{\supo}{SUPO\xspace}
\newcommand{\ci}[2]{{\scriptsize[#1, #2]}}
\definecolor{bestbg}{RGB}{198,219,239}
\definecolor{secondbg}{RGB}{221,234,247}

\title{Forget Freely, Grep Later:\\ Lossless, Index-Free Memory for Long-Horizon LLM Agents}

\usepackage{fontawesome5}

\newif\ifcameraready
\camerareadytrue          

\author{
  Ziyan Yang\thanks{Equal contribution.} \quad
  Taiwei Shi\footnotemark[1] \quad
  Jieyu Zhao \\
  University of Southern California \\
  \texttt{\{ziyany,taiweish,jieyuz\}@usc.edu}
  \ifcameraready
  \\[6pt]
  \href{https://github.com/limenlp/GrepMem}{\faGithub\ Code}\quad
  \href{https://grepmem.lovable.app/}{\faGlobe\ Interactive Demo}
  \fi
}

\usepackage{fvextra}
\usepackage[most]{tcolorbox}

\fvset{
  breaksymbolleft={},
  breaksymbolright={}
}

\newtcolorbox{promptbox}[1]{
  title={#1},
  colback=gray!5,
  colframe=gray!50!black,
  colbacktitle=gray!50!black,
  coltitle=white,
  fonttitle=\bfseries\normalsize,
  toptitle=4pt,
  bottomtitle=4pt,
  boxrule=0.5pt,
  arc=2pt,
  boxsep=0pt,
  left=5pt,
  right=5pt,
  top=5pt,
  bottom=5pt,
  breakable
}

\begin{document}
\maketitle

\input{sections/abstract}
\input{sections/intro}
\input{sections/related}
\input{sections/method}
\input{sections/setup}
\input{sections/results}
\input{sections/conclusion}

\bibliography{custom}

\appendix
\input{sections/appendix}

\end{document}

%% file: sections/abstract.tex
\begin{abstract}
Long-horizon LLM agents often exceed their context windows, forcing them to discard information that may be needed later. Existing context management strategies rely on summarization, masking, or structured memory, requiring decisions about what to preserve or how to retrieve it before future needs are known. However, these methods can introduce lossy compression, potentially discarding details needed for future tasks. To address this limitation, we introduce \method, a simple, lossless, and index-free memory mechanism built on two components: an append-only plain-text log of the agent's complete history and a regular-expression search tool for on-demand recall. Unlike structured memory systems, \method requires no embeddings, retrieval index, memory extraction, or additional models. We train Qwen3.5-9B with reinforcement learning under five context management strategies and evaluate on multi-question research sessions, long-stream recall, and agent-trajectory question answering.  \method achieves 95--97\% accuracy on \cb, versus at most 21\% for lossy baselines, and outperforms SUPO's summarization by 8.8 percentage points on BCP-Multi. \method scales effectively to increasingly long input streams while remaining on the accuracy–token-cost Pareto frontier across all five evaluation tasks. Our results show that a plain-text log and grep provide a simple but competitive approach to long-horizon agent memory, without requiring complex structured memory infrastructure.
\end{abstract}


%% file: sections/intro.tex
\section{Introduction}
\label{sec:intro}


\begin{figure}[t]
\centering
\includegraphics[width=\linewidth]{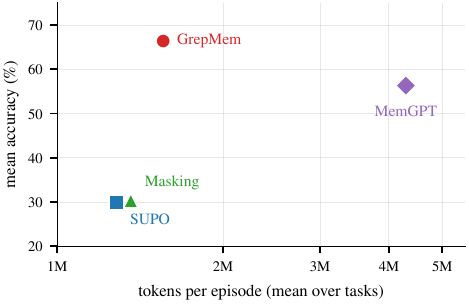}
\caption{\textbf{Mean Accuracy vs. Token Cost After RL.} We compare four context management strategies by their mean accuracy and token usage across five evaluation tasks. SUPO compresses context through summarization, while Masking discards older tool outputs to free context space. MemGPT maintains editable working memory, archival memory, and searchable message history. Our method, \method, combines summarization with a lossless, searchable history log, enabling on-demand retrieval through grep without a retrieval index. Token usage includes prompt and generated tokens across all model calls. Figure~\ref{fig:cost} reports per-task results.}
\label{fig:teaser}
\end{figure}

Long-horizon LLM agents accumulate interaction histories that eventually exceed their context windows, forcing them to make room for new observations while retaining useful information.
In deep research, for example, an agent may inspect many documents before combining its findings into a final answer \citep{wei2025browsecomp,chen2025browsecompplus,gao2025asearcher}.
An identifier or fact encountered early may become relevant only after several later searches, making it difficult to decide which details to retain when reducing context. Keeping the entire trajectory in context is expensive and eventually exceeds the window; even within that window, models do not reliably use all available information \citep{liu2024lost}.
Therefore, the challenge is not only to manage the working context but also to preserve the ability to recover previously observed information on demand.

Existing approaches address this challenge through compression or external memory but can incur additional computational overhead while risking the loss of important information needed for future tasks. Yet unlike human memory, digital storage allows agents to preserve complete interaction histories beyond the limited context window at relatively low cost. Recent studies further suggest that simple lexical search over raw histories can rival structured memory systems \citep{refind,grepallyouneed}, offering a lightweight way to recover past information on demand. Taking this idea to its logical extreme, we revisit the boundary between short-term context management and long-term memory in LLM agents: \emph{can a complete digital record and a simple search tool provide effective agent memory without a structured store or retrieval index?}

\begin{figure*}[!t]
\centering
\includegraphics[width=\textwidth]{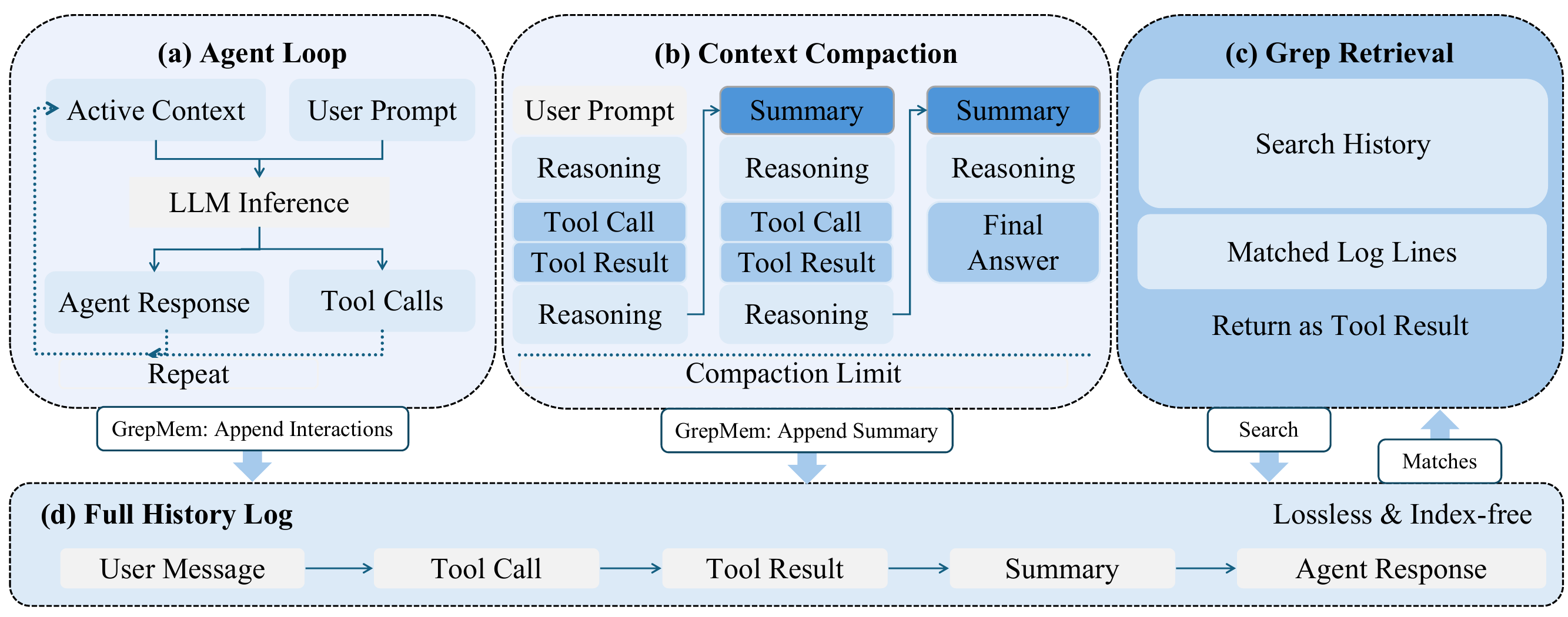}
\caption{\textbf{\method Overview.}
(a) The basic agent loop generates responses and tool calls from
its active context, while \method{} additionally appends these interactions and tool results to a persistent history log.
(b) Context compaction generates a summary when the working context reaches the compaction limit, allowing the agent to resume execution with a compacted context.  \method builds on \supo{}'s summarization loop.
Extending this framework, \method enables (c) on-demand retrieval through \grep and (d) a lossless, append-only full history log containing user messages, tool calls, tool results, summaries, and agent responses. By combining context summarization with a log preserving every message and tool output verbatim, \method enables \texttt{\grep} to retrieve matching log lines after the first compaction without embeddings, a retrieval index, or an additional model.}
\label{fig:overview}
\end{figure*}

We introduce \method, which separates the agent's bounded working context from its persistent history (Figure~\ref{fig:overview}).
We build on the summarization loop of \supo~\citep{lu2025supo}, a reinforcement learning (RL) algorithm that jointly optimizes tool use and context summarization. When the context reaches a length threshold, the agent summarizes its interaction history and resumes from the initial prompt and the summary.
\method extends this loop by automatically appending every message and tool output to a persistent plain-text log.
After the first compaction, the agent can retrieve earlier observations through a single tool, \grep, which searches the log using regular expressions and supports both
retrieving and counting matching lines. In \method, summaries provide context for ongoing reasoning, while the lossless log preserves all details omitted during compaction.
No embeddings, retrieval index, memory extraction, or additional models are required.

We train Qwen3.5-9B \citep{qwen35} under five context management strategies: ReAct \citep{yao2023react}, observation masking \citep{lindenbauer2025complexity}, \supo \citep{lu2025supo}, MemGPT-style memory \citep{packer2023memgpt}, and \method (Table~\ref{tab:managers}) using the same RL recipe, training data, and task tools (\S\ref{sec:setup}). We evaluate each strategy before and after RL training on multi-question research with \bcpm, constructed from \bcp question triples \citep{chen2025browsecompplus}; long-stream recall on \cb \citep{clm}; and agent-trajectory question answering on \ama \citep{zhao2026amabench}. Together, these settings test both retaining findings for known questions and recovering details for questions revealed later.

Overall, \method improves accuracy while remaining on the accuracy--token-cost Pareto frontier for all five evaluation tasks (\S\ref{sec:cost}, Figures~\ref{fig:teaser} and~\ref{fig:cost}). Its greatest advantage emerges when exact details must remain accessible beyond the active context window.  \method achieves 95--97\% accuracy on \cb, compared with at most 21\% for SUPO's summarization~\citep{lu2025supo} and observation masking \citep{lindenbauer2025complexity}. 
On \bcpm, it improves accuracy by 8.8 percentage points over \supo and 7.0 over MemGPT. These gains come at competitive token costs: on ContextBench, \method uses only 29\% as many tokens as MemGPT while achieving substantially higher accuracy on Log Triage.

Our contributions are threefold:
\begin{itemize}[leftmargin=*,itemsep=0pt,topsep=2pt,parsep=0pt]
  \item \textbf{Lossless memory without an index.} We introduce \method, which pairs lossy context summarization with a lossless trajectory log preserving the full interaction history, enabling the agent to recover omitted details through a single search tool, \texttt{grep\_history} (Figure~\ref{fig:overview}). 
  \item \textbf{Pareto-optimal accuracy--token-cost trade-offs.} We compare five context management strategies before and after RL training across three complementary settings.   \method lies on the accuracy–token-cost Pareto frontier across all five evaluation tasks, achieving higher accuracy than summarization-based baselines at competitive token costs (Figures~\ref{fig:teaser} and~\ref{fig:cost}, Table~\ref{tab:main}).  
  
  \item \textbf{Scalable and accurate recall over long histories.} \method maintains high accuracy on ContextBench's KV Store and Log Triage tasks  (Table~\ref{tab:benchmarks}) even as input streams grow far beyond the model's context window, enabling accurate and scalable long-horizon recall without complex memory infrastructure (Figure~\ref{fig:length}).
\end{itemize}

%% file: sections/related.tex
\section{Related Work}
\label{sec:related}

\begin{table*}[t]
\centering
\footnotesize
\setlength{\tabcolsep}{4pt}
\begin{tabular*}{\textwidth}{@{\extracolsep{\fill}}llcccc@{}}
\toprule
Method & How it works & \makecell{Scales past\\the window} & \makecell{Keeps\\everything} & Recall & No index \\
\midrule
ReAct \citep{yao2023react} & keeps all, stops when full & \xmark & \cmark & \xmark & \cmark \\
Masking \citep{lindenbauer2025complexity} & hides old tool outputs & \cmark & \xmark & \xmark & \cmark \\
\supo \citep{lu2025supo} & summarizes and restarts & \cmark & \xmark & \xmark & \cmark \\
MemGPT \citep{packer2023memgpt} & summarizes, manages memory & \cmark & \cmark & \cmark & \xmark \\
\midrule
\textbf{\method} (ours) & summarizes, greps raw log & \cmark & \cmark & \cmark & \cmark \\
\bottomrule
\end{tabular*}
\caption{\textbf{Comparison of Context Management Strategies.} We compare how five strategies manage context, preserve interaction history, and support retrieval.
\method augments \supo with an automatically appended plain-text history log and a grep tool, preserving the full history for retrieval without an index.}
\label{tab:managers}
\end{table*}

\paragraph{Compressing the working context.}
Context management strategies extend agent trajectories by selecting what remains in the active context.
Observation masking \citep{lindenbauer2025complexity} replaces older tool outputs with placeholders while preserving the agent's reasoning and actions, retaining recent observations within a fixed window, while ReSum \citep{wu2025resum} and \supo \citep{lu2025supo} use summaries to continue beyond the window.
Learned alternatives update bounded memory states (MemAgent and MEM1; \citealp{yu2025memagent,zhou2025mem1}), compress sub-trajectories (Context Folding and AgentFold; \citealp{sun2025contextfolding,ye2025agentfold}), or edit the context directly (Context Language Models; \citealp{clm}).
These mechanisms determine which details remain available in context. Following \supo{}’s summarization loop, \method separates working context compression from access to past experience: summaries maintain a compact working context, while the original history remains searchable so that omitted details can be recovered.

\paragraph{Organizing external memory.}
External stores preserve information beyond the working context, but differ in representation and access.
MemGPT \citep{packer2023memgpt} combines searchable message history with working and archival memory; Generative Agents \citep{park2023generative} retrieve recorded experiences by recency, importance, and relevance.
Other systems organize linked notes (A-MEM; \citealp{xu2025amem}), extracted facts (Mem0; \citealp{chhikara2025mem0}), or knowledge graphs (Zep and HippoRAG; \citealp{rasmussen2025zep,gutierrez2024hipporag}). These approaches introduce decisions about memory
organization and retrieval in addition to solving the task
\citep{yang2026selfevolvingllmmemoryextraction}.
LongMemEval \citep{wu2025longmemeval} and MemoryAgentBench \citep{hu2025memoryagentbench} also evaluate complementary aspects of long-term memory. Unlike prior approaches, \method retrieves information directly from a plain-text history log through a single index-free search tool. 


\paragraph{Searching history and learning to recall.}
Recursive Language Models \citep{zhang2025rlm} expose long inputs to programmatic inspection, and coding agents interact with files and environments through code and specialized tools \citep{yang2024sweagent,song2026coact}.
Closer to our setting, \citet{grepallyouneed} compare grep with vector retrieval over chat archives, while ReFind \citep{refind} uses agent-controlled lexical search over unmodified conversations.
These studies motivate raw-history retrieval but do not train agents to search their own growing trajectories.
RL has also been used to train agents for tool-based research \citep{jin2025searchr1,li2025websailor,wu2025webdancer,gao2025asearcher,shi2026experientialreinforcementlearning}.
Related work improves RL training efficiency through adaptive curricula \citep{shi2026efficient} and agent generalization through reusable skills extracted from successful trajectories \citep{xu-etal-2026-skill-reuse}.
We bring these directions together by training recall alongside task actions and summarization, and comparing five memory mechanisms.

%% file: sections/method.tex
\section{\method}
\label{sec:method}

\begin{algorithm}[t]
\caption{\method: one research episode}
\label{alg:grepmem}
\footnotesize
\begin{algorithmic}[1]
\State \textbf{Inputs:} policy $\pi_\theta$; task; interaction limits
\State \textbf{Initialize:} context $c$ and history log with the system prompt and task; disable grep
\While{interaction budget remains}
  \Statex \hspace{\algorithmicindent}\textcolor{blue!50!black}{\textit{// Context compaction}}
  \If{context approaches the window limit}
    \State Generate summary $s \sim \pi_\theta(\cdot\mid c,\text{summarize})$
    \State Append $s$ to the history log
    \State Reset $c$ to the system prompt, task and $s$
    \State Enable \grep
  \EndIf
  \Statex \hspace{\algorithmicindent}\textcolor{blue!50!black}{\textit{// Act or recall}}
  \State Sample reply $a \sim \pi_\theta(\cdot\mid c)$
  \State Append $a$ to context $c$ and the history log
  \If{$a$ contains a final answer}
    \State \Return the answer in $a$
  \EndIf
  \State Execute the requested tools; grep reads the log
  \State Append tool outputs to $c$ and the history log
  \State Update interaction budgets
\EndWhile
\State Ask $\pi_\theta$ for a final answer from $c$, without tools
\State \Return the answer
\end{algorithmic}
\end{algorithm}

\subsection{Bounded Context, Persistent History}

As discussed in the introduction, memory is essential for long-horizon agents, yet existing approaches face a trade-off: context compression risks losing information beyond the context window, while long-term memory systems often introduce additional complexity in memory retrieval. These limitations highlight the potential of combining context compaction with simple, long-term access to the complete interaction history to address both information loss and retrieval complexity.

\method revisits the relationship between short-term context management and long-term memory by maintaining a bounded working context alongside a lossless interaction history.  Algorithm~\ref{alg:grepmem} summarizes the overall procedure in detail. Building on the summarization loop of \supo{}~\citep{lu2025supo}, our context compaction mechanism generates a
summary when the working context approaches its limit, allowing the agent to resume from the original prompt and summary.
Meanwhile, an append-only log preserves all interactions, including messages,
tool calls, outputs, and summaries, making the interaction history \emph{lossless}: nothing
is discarded, even after the working context is compacted.
Together with on-demand grep retrieval,
this gives the agent a compact working context and a complete, searchable
history (Figure~\ref{fig:overview}). During an episode, the agent iteratively generates responses and invokes task-specific tools to interact with its environment. Each tool call returns an observation to the working context, while the harness appends the agent's messages, tool calls, and outputs to the persistent history log. These two representations serve complementary purposes: the working context supports ongoing reasoning through summarization, while the lossless history enables on-demand recovery of details omitted during compaction. Appendix~\ref{app:impl} describes how  summarization input fits within the context window.

\subsection{On-Demand Recall with Grep}
After the first compaction, \grep becomes available during subsequent task execution. The agent decides when to invoke it, subject to the interaction budget. The tool searches the persistent history log using case-insensitive regular expressions. By default, it returns matching lines in chronological order, with surrounding context and their original locations in the history, enabling recovery of details omitted during compaction. 

To limit response size, results are paginated, and the agent can request additional pages to access the remaining matches. The tool also supports a count-only mode that reports the total number of matching lines without returning their contents. Because retrieval relies on regular-expression matching rather than semantic similarity, the agent must formulate search patterns that match relevant text in the history log. These patterns may be simple keywords, such as names or identifiers, or more complex regular expressions. Appendices~\ref{app:prompts} and~\ref{app:impl} give the tool schema, output limits, and examples.




\subsection{Learning with Reinforcement Learning}
Since \method preserves complete histories and enables on-demand retrieval, effective memory use requires the agent to decide when and how to retrieve relevant information. We therefore use reinforcement learning to teach agents these capabilities by jointly optimizing task actions, context summarization, and memory retrieval through a shared episode-level reward, without separate supervision for summaries or grep queries. We evaluate \method both before and after RL training, with details provided in
Section~\ref{sec:setup} and Table~\ref{tab:main}.


%% file: sections/setup.tex
\section{Experimental Setup}
\label{sec:setup}

We evaluate \method{} with four baselines across five tasks spanning multi-question research, recorded-trajectory analysis, and recall over long input streams. \method{} achieves the highest average accuracy both before and after RL training (Table~\ref{tab:main}) and lies on the accuracy--token-cost Pareto frontier across all five tasks (\S\ref{sec:cost}).

\subsection{Context Management Strategies}
We compare five context management strategies, summarized in Table~\ref{tab:managers}.
\textbf{ReAct} \citep{yao2023react} retains full history until the context limit is reached. 
\textbf{Masking} \citep{lindenbauer2025complexity} replaces older tool observations with placeholders.
\textbf{\supo} \citep{lu2025supo} summarizes and resumes the context.
\textbf{MemGPT-style memory} \citep{packer2023memgpt} combines a notepad with searchable message history and archival memory.
\textbf{\method} augments \supo with a lossless, append-only history log and an index-free grep interface as described in \S\ref{sec:method}. Implementation details appear in Appendices~\ref{app:impl} and~\ref{app:baselines}.


\subsection{Training and Shared Settings}

To ensure a fair comparison, we separately train Qwen3.5-9B~\citep{qwen35} with thinking enabled under each context management strategy using the same GRPO-based RL recipe~\citep{shao2024deepseekmath}. All strategies use a 32,768-token context window and the same interaction budgets.

Our training set contains 4,000 episodes: 70\% single-question OpenResearcher tasks~\citep{li2026openresearcher}, 15\% three-question sessions constructed from the OpenResearcher training split, and 15\% synthetic ContextBench-style streams~\citep{clm}. This mixture targets complementary capabilities: research and tool use in single-question tasks, retention of findings across questions in multi-question sessions, and precise lookup and counting over long histories in synthetic streams. OpenResearcher tasks use binary correctness rewards, averaged across questions in multi-question sessions. Details on optimization, data construction, and interaction budgets appear in Appendix~\ref{app:training}. 

\begin{table}[t]
\centering
\small
\setlength{\tabcolsep}{4pt}
\begin{tabular}{@{}lrc@{}}
\toprule
Task & Questions & Timing \\
\midrule
\bcpm & 828 & before \\
\ama file & 2{,}496 & before \\
\ama stream & 2{,}268 & after \\
\cb KV & 576 & after \\
\cb Log & 480 & after \\
\bottomrule
\end{tabular}
\caption{\textbf{Evaluation Task Sizes.} Questions are totals per evaluation run. Timing indicates whether questions are revealed before or after the relevant input.}
\label{tab:benchmarks}
\end{table}

\subsection{Evaluation Tasks}
We evaluate \method on \textbf{\bcpm}, \textbf{AMA-Bench file and stream} \citep{zhao2026amabench}, and \textbf{ContextBench} \citep{clm}, as summarized in Table~\ref{tab:benchmarks}. These tasks are distinguished by whether questions are revealed before or after the relevant input. When questions are known in advance, agents can use them to guide retention and search. This setting tests whether agents can retain findings while investigating multiple research questions (BCP-Multi) or search recorded
trajectories with a shared search tool (AMA-Bench file). When questions are revealed after the input, agents must process the input without knowing future needs, so answering depends on whether relevant details remain available in the working context or, for strategies supporting retrieval, through memory search (AMA-Bench stream and ContextBench).

\paragraph{Questions provided before the input.} 
\textbf{\bcpm} extends single-question evaluation to test retention across research questions. We group three questions into one episode so that findings from earlier searches must remain accessible while the agent investigates other questions and prepares a joint final answer. 
We construct it from \bcp \citep{chen2025browsecompplus} by shuffling its 830 BrowseComp questions \citep{wei2025browsecomp} with a fixed seed into 276 disjoint triples, omitting the remaining two questions. 
All three questions are provided together in the initial prompt and are answered together within 60 turns over the original 100K-document corpus. The agent retains its context across all three questions in each episode. \textbf{\ama file} uses recorded agent trajectories and associated question–answer pairs provided by the benchmark \citep{zhao2026amabench}, with the source trajectories kept outside the model's working context and accessible through \texttt{grep\_trajectory}. We evaluate 208 trajectories, with the agent answering all 12 questions for each trajectory in a single session with a shared budget of 60 turns. We provide all context management strategies with \texttt{grep\_trajectory}, a tool that searches recorded actions and observations using regular expressions and returns matching lines grouped by step. Unlike \texttt{grep\_history}, which searches the agent's own interaction history, \texttt{grep\_trajectory} searches the benchmark-provided fixed source trajectory.

\paragraph{Questions revealed after the input.}
\textbf{\cb} \citep{clm} evaluates whether agents can recall information from long sequences of key–value records or service logs. The agent receives these records incrementally, then answers 24 questions about them. KV Store requires retrieving exact values, while Log Triage requires finding specific log entries and counting entries that satisfy given conditions. We evaluate input sequences ranging from 33K to 968K tokens across 11 length levels, with four random seeds per level.
\textbf{\ama stream} evaluates question answering over recorded agent trajectories presented incrementally. Unlike AMA-Bench file, which stores the source trajectory as a searchable file, AMA-Bench stream feeds each trajectory into the agent’s context in successive chunks of approximately 12K characters. We evaluate 189 trajectories, each containing at most 128K tokens. After the agent has processed an entire trajectory, we save its final context and memory state. Each of the trajectory’s 12 questions is then answered independently, starting from an independent copy of the final context and memory state. Both settings hide questions during input processing, preventing question-guided retention.

\subsection{Metrics and Reporting}
We evaluate the model before and after RL training. OpenResearcher, BCP-Multi, and AMA-Bench answers are graded by gpt-5.6-sol~\citep{openai2026gpt56sol} against references; ContextBench uses its official programmatic graders. We report 95\% confidence intervals for accuracy. These intervals reflect variation across test items, not across training runs. We report total token cost across all model calls, including summarization and recall (Appendix~\ref{app:eval}).

%% file: sections/results.tex
\section{Results and Analysis}
\label{sec:results}

\begin{table*}[t]
\centering
\small
\setlength{\tabcolsep}{5pt}
\resizebox{\textwidth}{!}{%
\begin{tabular}{@{}l cc c ccc c c@{}}
\toprule
& \multicolumn{2}{c}{Questions known before the input} && \multicolumn{3}{c}{Questions revealed after the input} && \\
\cmidrule{2-3} \cmidrule{5-7}
Strategy & \bcpm & \ama file && \ama stream & \cb KV & \cb Log && Avg. \\
\midrule
\multicolumn{9}{@{}l}{\textit{After RL}} \\
ReAct          & 0.2 & 38.3 && 26.6 & 0.0 & 0.0 && 13.0 \\
\supo          & 38.1 & 45.5 && 45.2 & 13.9 & 6.6 && 29.8 \\
Masking        & 34.8 & 44.5 && 44.9 & 20.7 & 6.6 && 30.3 \\
MemGPT         & \underline{39.9} & \textbf{46.4} && \underline{45.4} & \textbf{99.9} & \underline{50.0} && \underline{56.3} \\
\method        & \textbf{46.9} & \underline{46.3} && \textbf{46.1} & \underline{97.4} & \textbf{95.3} && \textbf{66.4} \\
\midrule
\multicolumn{9}{@{}l}{\textit{Before RL}} \\
ReAct          & 0.0 & 22.4 && 26.2 & 0.0 & 0.0 && 9.7 \\
\supo          & \textbf{45.2} & \underline{45.5} && 41.7 & 10.9 & 6.6 && 30.0 \\
Masking        & 18.6 & 42.3 && 42.1 & 13.3 & 5.5 && 24.4 \\
MemGPT         & 34.4 & 45.4 && \textbf{46.1} & \textbf{95.8} & \underline{48.9} && \underline{54.1} \\
\method        & \underline{40.9} & \textbf{46.9} && \underline{45.8} & \underline{93.9} & \textbf{75.0} && \textbf{60.5} \\
\bottomrule
\end{tabular}}
\caption{\textbf{Accuracy Before and After RL Training.}
We compare the accuracy (\%) of five context management
strategies using Qwen3.5-9B before and after training with
the same RL recipe (\S\ref{sec:setup}).
Within each block, \textbf{bold} marks the best value
in each column, and \underline{underlining} marks the second best.
For \cb, we report exact-value accuracy on KV Store
and exact-answer accuracy on Log Triage.
Avg.\ denotes the mean accuracy across five tasks.}
\label{tab:main}
\end{table*}

\begin{figure*}[t]
\centering
\includegraphics[width=\textwidth]{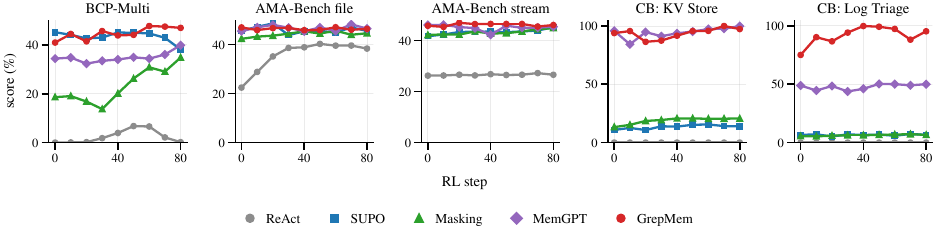}
\caption{\textbf{Evaluation Performance of Five Context Management Strategies During RL Training.} Scores are averaged over the evaluation samples available at each checkpoint. Evaluation details are provided in Appendix~\ref{app:eval}.}
\label{fig:curves}
\end{figure*}

We evaluate the effectiveness of different context management strategies in terms of accuracy, token cost, and how their performance changes with RL training. Table~\ref{tab:main} compares accuracy before and after RL, while Figure~\ref{fig:curves} tracks performance throughout training. Building on these results, we further examine when \method improves accuracy, how RL affects different context management strategies, how performance varies with stream length, and the trade-off between accuracy and token cost.

\subsection{Where Does \method Improve Accuracy?}
\label{sec:main}

\paragraph{Questions known before the input.}
On \bcpm, the RL-trained \method achieves 46.9\% accuracy, outperforming \supo by 8.8 percentage points and MemGPT by 7.0 points. Although all three questions are provided in the initial prompt, the agent must retain findings from earlier searches while investigating subsequent questions.
\supo and \method average 25--27 context compactions per episode, testing whether these findings remain accessible after repeated summarization. \method can recover findings omitted during compaction by searching its interaction history.
On \ama file, by contrast, all strategies can revisit the original trajectory through a shared search tool, while all strategies except ReAct perform similarly, suggesting that searching the agent's own history provides the greatest benefit when the original information is no longer directly accessible.

\paragraph{Questions revealed after the input.}
The largest gains over lossy baselines occur on \cb, where KV Store and Log Triage require exact values or individual log entries that summarization can discard, making them well suited to lossless recall.
After RL, \method achieves 97.4\% accuracy on KV Store and 95.3\% on Log Triage, compared with just 6.6--20.7\% for \supo and Masking. However, revealing questions after the input does not by itself make lossless recall beneficial.
On \ama stream, all strategies except ReAct differ in accuracy by at most 1.2 percentage points after RL, and \method's lead over \supo narrows from 4.1 points before RL to 0.9 points after RL.
Even on trajectories exceeding the context window, the gain is not statistically significant. Unlike \cb, many \ama stream questions concern causes or state changes that summaries can preserve, rather than exact values or log entries. These results suggest that the value of lossless recall depends more on whether a question requires details omitted by summarization than on when the question is revealed.

\input{sections/analysis}
\subsection{How Does RL Affect Context Management Strategies?} 
\label{sec:length}

\begin{figure}[t]
\centering
\includegraphics[width=\linewidth]{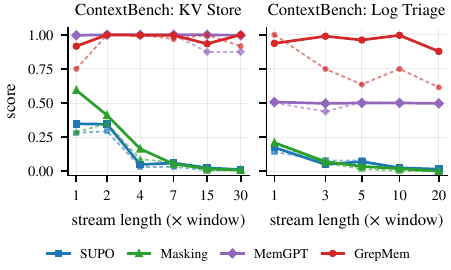}
\caption{\textbf{ContextBench Performance Across Stream Lengths.} We evaluate how stream length affects ContextBench performance before (dashed) and after RL (solid) training. }
\label{fig:length}
\end{figure}

\begin{figure*}[t]
\centering
\includegraphics[width=\textwidth]{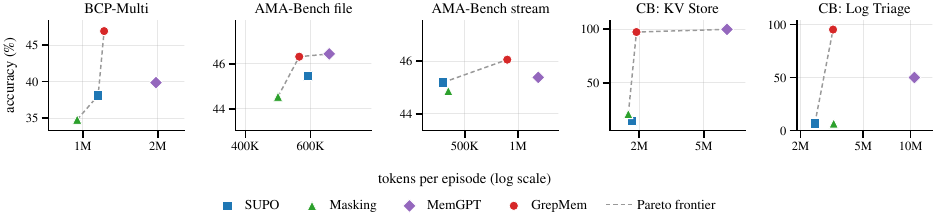}
\caption{\textbf{Per-Task Accuracy vs. Token Cost After RL.} We compare the accuracy and token usage of RL-trained strategies on each evaluation task. Cost is the number of tokens processed per episode (prompt and generated tokens over all LLM calls; \cb: per stream), on a log scale. Dashed: the Pareto frontier.}
\label{fig:cost}
\end{figure*}

As shown in Table~\ref{tab:main}, RL improves the average accuracy of most context management strategies, with \method{} gaining 5.9 percentage points to achieve the highest post-RL average accuracy of 66.4\%. The largest improvement occurs on \textbf{ContextBench Log Triage}, where accuracy increases by 20.3 points
since it requires lookup and counting over service logs. RL may improve query formulation and evidence aggregation, enabling \method to retrieve and utilize historical information more effectively.  On \textbf{\cb KV Store}, accuracy increases by 3.5 points.
Since this task requires exact-value
retrieval for a given key, RL may help \method formulate
more effective search queries.
On \textbf{BCP-Multi}, accuracy improves by 6.0 points.
As agents must retain and reuse
findings across multiple questions in a shared session,
RL may improve search timing, query formulation,
and evidence reuse. In contrast, \textbf{AMA-Bench file and stream} show
almost no improvement after RL (-0.6 and +0.3 points,
respectively). These tasks require reasoning about
state transitions and causal relationships across
agent trajectories, which retrieval optimization alone may not sufficiently address.

These results suggest that RL gains depend not only on whether historical information remains accessible, but also on whether the task's required operations can be improved through optimizing the agent's tool-use and evidence-utilization behavior. 





\subsection{How Does Performance Scale with Stream Length?}
As shown in Figure~\ref{fig:length}, \method maintains strong performance on both KV Store and Log Triage as stream length increases, demonstrating its ability to scale to longer input streams. In contrast, \supo and masking exhibit substantial performance degradation as stream length grows, with accuracy approaching zero in some settings. MemGPT maintains relatively stable performance on Log Triage but remains substantially below \method, with accuracy around 50\%. These widening gaps suggest that preserving searchable history becomes increasingly valuable as input length grows.
Furthermore, RL improves \method's performance on Log Triage at longer stream lengths, while its KV Store accuracy remains high both before and after training. These results suggest that \method's searchable history enables effective scaling to longer input streams by preserving access to relevant information, while RL further improves evidence utilization. Such capabilities are valuable for long-running agents, where historical information accumulates beyond the model's context window.



\subsection{What Does Grep Cost?}
\label{sec:cost}

\method lies on the accuracy--token-cost Pareto frontier across all five benchmark settings: no evaluated strategy is both cheaper and more accurate (Figures~\ref{fig:teaser} and \ref{fig:cost}). 
Compared with \supo, it uses only 6\% more tokens on BCP-Multi and 5\% fewer on AMA-Bench file while achieving higher accuracy. On ContextBench, it substantially outperforms \supo with only 19\% additional token usage and consumes only 29\% as many tokens as MemGPT. The notable exception is AMA-Bench stream, where \method uses 2.3× as many tokens as \supo for a small improvement in accuracy and calls grep roughly 19 times per trajectory.  These results suggest that recall incurs additional retrieval cost, but it is worthwhile and most cost-effective when the retrieved evidence materially improves the final answer.

%% file: sections/conclusion.tex
\section{Conclusion} 
\label{sec:conclusion}

We introduce \method, a simple, lossless, and index-free memory mechanism that combines context summarization with a complete, searchable interaction history. Our results demonstrate that \method supports accurate long-horizon recall, scales effectively to increasingly long input streams, and lies on the accuracy–token-cost Pareto frontier across all five evaluation tasks. Comparisons before and after RL training further show that RL can improve how agents retrieve and utilize historical information, with benefits varying across tasks and context management strategies. 

Our work reexamines the boundary between short-term context management and long-term memory in LLM agents. What should an agent forget when it can remember everything, and how much complexity does remembering truly require? 
By preserving complete histories alongside compact working contexts, \method suggests that lossless preservation and lossy compaction can serve complementary purposes. Looking ahead, we envision \method as a simple but competitive baseline for evaluating future methods, inspiring principled memory designs for reliable long-running agents.

\section*{Limitations}
Our evaluation focuses on controlled benchmarks and a single model, leaving open how \method generalizes to broader agent settings. However, these benchmarks cover diverse memory requirements, from multi-question research to exact recall over long histories, providing evidence for the benefits of lossless memory. \method also relies on lexical search, which may be less effective when relevant information cannot be identified through explicit search patterns. Nevertheless, our goal is not to replace more sophisticated memory systems in every setting, but to demonstrate that preserving raw history and retrieving it with a simple tool can already provide strong performance without structured memory infrastructure. Finally, the benefits of lossless recall depend on the task: when summaries retain sufficient information, additional retrieval may introduce unnecessary overhead. This suggests that an important future direction is to determine when simple, lossless retrieval is most valuable and how to use it selectively.

\section*{Ethics Statement}
Our experiments use established benchmarks, fixed offline corpora, and publicly available agent trajectories, without involving human subjects or collecting new personal data. \method preserves complete interaction histories, which may contain sensitive information in real-world deployments. This concern is shared by other memory systems that retain historical interactions, rather than being unique to our approach. As with any persistent memory mechanism, appropriate access controls and data retention policies should be applied according to the deployment setting. Our work focuses on improving long-horizon agent reliability through a simple and transparent memory mechanism. To support transparency and reproducibility, we will release our code, prompts, and trained model checkpoints.

%% file: sections/appendix.tex
\newpage
\section{Prompts}
\label{app:prompts}

All prompts are copied from the harness. The placeholder \texttt{\{extra\_tools\}} is empty until the first compaction and then holds the \grep line below.
\begin{promptbox}{System Prompt (Deep Research)}
\begin{Verbatim}[breaklines,breakanywhere,fontsize=\normalsize]
You are a deep-research agent. Answer the user's question by researching a fixed document corpus with the tools provided.

Tools:
- search(query): dense retrieval over the corpus; returns the top 10 documents (docid, url, first ~512 tokens).
- get_document(docid): returns the full text of a document (up to 8192 tokens).
{extra_tools}
Guidelines:
- Questions often require combining several clues; break them down, search iteratively, and verify candidates against the evidence.
- Prefer short, specific search queries. Open documents when snippets are insufficient.
- When you are confident, stop calling tools and give your final answer as a short phrase inside \boxed{}, e.g. \boxed{Marie Curie}.
- You have at most 30 turns (assistant replies). The number of turns left is shown after each tool result; give your final answer before it runs out.
\end{Verbatim}
\end{promptbox}
ReAct omits the last guideline.
\method adds ``Replies that only call grep\_history do not count as turns (up to 10 such replies).'' MemGPT adds an equivalent instruction for its memory tools.

\begin{promptbox}{\grep Description (After First Compaction)}
\begin{Verbatim}[breaklines,breakanywhere,fontsize=\normalsize]
- grep_history(pattern, max_matches, offset, context_lines, count): regex search over your full, uncompressed interaction history (all earlier tool outputs and messages), including parts removed from your context by summarization. Each match is labeled with the turn and tool call it came from.
\end{Verbatim}
\end{promptbox}
The tool's JSON schema documents the arguments (\texttt{max\_matches}: default 10, at most 30; \texttt{offset}: matches to skip; \texttt{context\_lines}: default 1, at most 3; \texttt{count}: only count matching lines). It recommends patterns made of previously observed exact strings (names, numbers, dates, IDs).

\begin{promptbox}{Compaction Prompt (\supo and \method)}
\begin{Verbatim}[breaklines,breakanywhere,fontsize=\normalsize]
Your context window is full and will be reset. Write a summary that lets you continue the research from scratch with only the original question and this summary. Include:
1. Key findings so far, with the docids/urls that support them.
2. Candidate answers and how well each is verified.
3. Clues already resolved and approaches/queries that did not work.
4. Concrete next steps.
Do not call any tools. Respond only with the summary.
\end{Verbatim}
\end{promptbox}
Stream tasks use the same prompt with ``the task instructions'' in place of ``the original question''.


\begin{promptbox}{Post-Compaction Message}
\begin{Verbatim}[breaklines,breakanywhere,fontsize=\normalsize]
{question}

---
Your earlier context was compacted. Summary of your progress so far:

{summary}

(The full raw history of the earlier context is searchable with grep_history.)

[{n} of 30 turns left]

Continue the research.
\end{Verbatim}
\end{promptbox}
The note in parentheses appears only for \method.

\paragraph{Masking and MemGPT.}
Masking \citep{lindenbauer2025complexity} adds to the system prompt: ``Your context window is limited: when it fills up, the outputs of your oldest tool calls are replaced by a short placeholder (your own messages and tool calls are kept).''
A masked output reads \texttt{[Old tool output removed from the context (n characters).]}.
MemGPT \citep{packer2023memgpt} adds a description of its context (``when it fills up, the oldest messages are removed from it and replaced by a summary (you are warned shortly before). Removed messages remain searchable with conversation\_search; keep what you need in your working context or archival memory.'') and of its memory tools:
\begin{promptbox}{MemGPT Memory Tools}
\begin{Verbatim}[breaklines,breakanywhere,fontsize=\normalsize]
- working_context_append(content) / working_context_replace(old_content, new_content): edit your working context, a notepad of up to 4000 characters shown at the end of this system prompt; it is never removed from your context.
- archival_memory_insert(content) / archival_memory_search(query, page): store passages in your unlimited archival memory and search them (keyword search).
- conversation_search(query, page): case-insensitive text search over every earlier message and tool output, including those removed from your context.
\end{Verbatim}
\end{promptbox}
\begin{promptbox}{MemGPT Memory Warning}
\begin{Verbatim}[breaklines,breakanywhere,fontsize=\normalsize]
[System message] Memory pressure: your context is about {pct}% full. When it is full, the oldest messages will be removed from it and replaced by a summary. Save anything you will need later to your working context or archival memory now.

Your context window is full. The messages above, after the task, will be removed from your context (they remain searchable with conversation_search); the most recent messages are kept. Write a summary of the messages that will be removed, so that you can continue the task with your working context, this summary and the kept messages. If the messages above begin with an earlier summary, merge it into the new one. Keep the key facts, findings and progress, copied exactly where later steps may need them.
Do not call any tools. Respond only with the summary.
\end{Verbatim}
\end{promptbox}

\paragraph{Turn budget and forced answer.}
After each tool result, strategies other than ReAct see \texttt{[\{n\} of 30 turns left]}; before the last turn the note reads ``this is your last turn. You must give your final answer in \verb|\boxed{}| now; do not call any more tools.''
An episode that runs out of turns (or whose final reply is cut off) gets one call without tools:
\begin{promptbox}{Forced-Answer Prompt}
\begin{Verbatim}[breaklines,breakanywhere,fontsize=\normalsize]
You have used all your turns. Do not call any more tools. Based on what you have found so far, give your final answer now as a short phrase inside \boxed{}; give your best guess if you are unsure.
\end{Verbatim}
\end{promptbox}
Three-question sessions ask for one numbered \verb|\boxed{}| line per question instead.
ReAct ends unanswered when its window is full, since there is no room for another call.

\section{Implementation Details}
\label{app:impl}

\paragraph{Log format and grep output.}
Each log record is a header line (e.g.\ \texttt{[turn 4 tool get\_document(...)]}) followed by its text, and \grep matches the text line by line; records of earlier \grep outputs are skipped, since they only repeat older matches.
The output starts with the number of matching lines by source, then lists the matches grouped under their record's header, with line numbers within the record (\texttt{:} marks a match, \texttt{-} a context line, \texttt{--} a gap).
Matches are returned oldest first, ten per page by default (at most 30), with one surrounding line by default (at most three).
Each response is capped at 6K characters, ends at a whole match and gives the offset for continuing the listing.
Lines are clipped to 300 characters around the match, an invalid pattern returns the compiler's error, and a search that takes more than 10\,s is stopped with an error.
An example on a small log:
\begin{promptbox}{Example of History Search and Counting}
\begin{Verbatim}[breaklines,breakanywhere,fontsize=\normalsize]
grep_history(pattern="doc_8812|born in 19\d\d", context_lines=1)

4 matching line(s) in 3 history entries (3 in tool outputs, 1 in summaries).
== [turn 3 tool search({"query": "Georgian composer opera 1960s"})]
  1: [1] docid: doc_8812
  2- url: https://example.org/composers/a
== [turn 4 tool get_document({"docid": "doc_8812"})]
  1: docid: doc_8812
  2- url: https://example.org/composers/a
  --
  4- Early life
  5: The composer was born in 1931 in Tbilisi and
  6- studied at the Tbilisi State Conservatory until 1954.
== [compaction 1 summary]
  1: Candidate composer found in doc_8812; birth year and conservatory still to verify.
(showing matches 1-4 of 4)

grep_history(pattern="doc_8812", count=true)

3 matching line(s) in 3 history entries (2 in tool outputs, 1 in summaries).
\end{Verbatim}
\end{promptbox}

\paragraph{Fitting the compaction call.}
Several large tool outputs in one turn can push the context past $\tau$ before the agent can compact, but the compaction prompt must still fit in $W$.
We estimate tokens from characters, calibrated on the last call's exact prompt-token count, and drop the oldest turns (always keeping the system prompt, the task and the latest turn), then shorten the latest tool outputs if needed, until the prompt fits.
If the model answers the compaction request with a tool call instead of text, its final reasoning is used as the summary.

\section{Baselines}
\label{app:baselines}

\paragraph{Masking.}
\citet{lindenbauer2025complexity} keep the last $M$ observations.
With a fixed context window, we mask observations according to a token budget. When the next model call would exceed $\tau$, the oldest unmasked observations are replaced by the placeholder until the context is below $\tau/2$. In stream tasks, these observations are the incoming user messages.
Since Qwen3.5 \citep{qwen35} retains the reasoning from
every turn in the current episode, masking observations
alone may not free enough space. In that case, we remove
reasoning from the oldest turns while preserving their
actions. We drop entire turns only if the prompt still
does not fit within the context window. As with compaction,
masking in batches keeps the prompt prefix unchanged
between masking events.

\paragraph{MemGPT.}
The main context is the system prompt with the working-context block, followed by a message queue headed by the task and a recursive summary.
At 70\% of $\tau$ the agent is warned.
At $\tau$, the queue is flushed: the oldest messages are evicted until about half the window remains, and the model writes a new recursive summary of the old summary and the evicted messages, in a call without tools.
Like MemGPT's summarizer, tools are disabled during this call. When tools are available, the base model returned a memory-tool call instead of a summary in about one-third of context flushes.
\texttt{conversation\_search} finds the messages (whole tool outputs or streamed batches) that contain a case-insensitive substring and reports their number, five per page.
Archival memory ranks passages with BM25 \citep{robertson2009bm25} instead of MemGPT's embeddings and also returns five per page.
On \cb, each streamed operation allows up to 8 memory-tool replies, and the chain of replies within one operation is trimmed if it would overflow the window.

\section{Training Details}
\label{app:training}

\paragraph{Data.}

We train on 4,000 episodes comprising three task types:
70\% single-question OpenResearcher episodes,
15\% three-question OpenResearcher sessions,
and 15\% synthetic ContextBench-style streams. OpenResearcher \citep{li2026openresearcher} questions are English questions with answers of at most ten words (3{,}013 questions), searched over the OpenResearcher corpus of 14.9M web documents with Qwen3-Embedding-8B \citep{zhang2025qwen3embedding}; \texttt{search} returns the top 10 documents with their id, URL and first ${\sim}$512 tokens.
\texttt{get\_document} returns up to 8{,}192 tokens of a selected document.
To construct three-question training sessions, we shuffle the OpenResearcher training split with seed 2, partition it into triples and take the first 600 triples.
Each triple is presented as three numbered questions in one initial prompt and answered together as numbered \verb|\boxed{}| lines.
The 60-turn budget is shared across the three questions, not enforced as separate 20-turn sessions.
A question may appear both in a single-question training episode and in a three-question session.
\cb training streams use seeds from 1000 on (evaluation uses 42--45), with KV Store at 1{,}000 or 2{,}000 records and Log Triage at 1{,}600 or 3{,}200 lines (1.5--6$\times$ the window), and 12 queries each.

\paragraph{Turns as training sequences.}
Each model call is one step.
Consecutive turns that extend the same context are merged into one training sequence. A context compaction or a new user turn in a stream starts a new sequence whose prompt is the full context available at that point.
Compaction and memory calls are sequences of their own, so summaries are trained too.
All training sequences within an episode share the same episode-level advantage, allowing RL to assign credit across context compactions without separate rewards for summaries or individual grep calls.

\paragraph{Hyperparameters and compute.}
Table~\ref{tab:hparams} lists the hyperparameters, which are shared by all strategies.
We train with rLLM \citep{rllm2025} on verl \citep{sheng2024hybridflow}, with vLLM \citep{kwon2023vllm} for rollouts.
Each run uses 24 H100 GPUs (8 for the FSDP trainer, 16 for rollouts; ReAct uses 8 GPUs for rollouts because its rollouts are shorter), and a dense-retrieval server on 3 more GPUs is shared by all runs.
In total, training used about 6K H100-hours; MemGPT's runs were the slowest, since its \cb rollouts make several hundred calls.

\begin{table}[ht]
\centering
\small
\resizebox{\linewidth}{!}{%
\begin{tabular}{@{}ll@{}}
\toprule
Base model & Qwen3.5-9B (thinking) \\
Algorithm & GRPO, std-normalized advantages \\
Prompts per step / rollouts per prompt & 8 / 8 \\
Steps & 80 \\
Learning rate / weight decay & $10^{-6}$ / 0.1 \\
Clip range (low / high) & 0.2 / 0.28 \\
KL penalty & low-variance KL, $\beta = 0.01$ \\
Loss aggregation & token mean \\
Dynamic sampling & drop uniform-reward groups \\
Off-policy correction & token-level TIS, cap 2 \\
Async staleness threshold & 0.5, partial rollouts \\
Temperature / top-$p$ & 0.7 / 0.95 \\
Window $W$ / generation $G$ / threshold $\tau$ & 32{,}768 / 4{,}096 / 28{,}672 \\
Turn budget $T$ & 30 per question \\
Free grep / memory-tool replies & up to 10 per episode \\
Reward & LLM judge (questions), grader (streams) \\
Retriever & Qwen3-Embedding-8B, top 10 \\
Hardware & 24$\times$H100 per run (8 train, 16 rollout) \\
\bottomrule
\end{tabular}}
\caption{\textbf{Training Configuration.} We report the model, optimization settings, interaction budgets, and hardware used for training. Training hyperparameters are shared across all context management strategies.}
\label{tab:hparams}
\end{table}

\begin{figure*}[t]
\centering
\includegraphics[width=\textwidth]{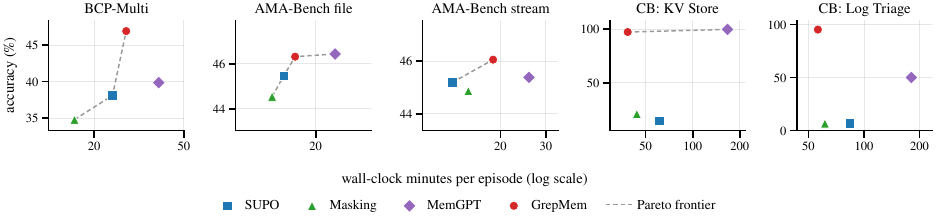}
\caption{\textbf{Per-Task Accuracy vs. Runtime After RL.} We compare the accuracy and wall-clock runtime of RL-trained strategies on each evaluation task. Runtime is measured in minutes per episode, or per stream for ContextBench, and is shown on a logarithmic scale. Dashed lines indicate the Pareto frontier. Runtime measurements are approximate because episodes ran concurrently on shared inference and retrieval servers.}
\label{fig:costtime}
\end{figure*}

\paragraph{Rewards.}
OpenResearcher tasks use binary correctness rewards,
averaged across questions in three-question sessions.
ContextBench uses its official programmatic grader.
Neither summarization nor individual grep calls receive separate rewards.

\section{Evaluation Details}
\label{app:eval}

\paragraph{Benchmarks.}
\bcpm uses the same tools and retriever as training, pointed at the \bcp \citep{chen2025browsecompplus} corpus.
We shuffle its 830 questions with Python's \texttt{random.Random(1).shuffle}, take consecutive non-overlapping triples and discard the final two questions. This produces 276 episodes containing 828 questions, grouped identically across strategies and checkpoints.
Each initial prompt presents three questions and requests numbered final answers. The agent retains its context and memory as it works through the questions.
\cb \citep{clm} uses six KV Store levels (R520–R16000) and five Log Triage levels (L640–L12800), with four random seeds per level. Operations are delivered one per turn, using the benchmark’s official 32K-token budget and graders. KV values end with a hash that the grader requires.
\ama \citep{zhao2026amabench} stream feeds each trajectory in ${\sim}$12K-character chunks (189 trajectories of at most 128K tokens, 2{,}268 questions) and asks each question on a copy of the final state, following the benchmark's per-query protocol.
\ama file gives every strategy \texttt{grep\_trajectory}, the same tool as \grep applied to the trajectory's steps. Each session answers all 12 questions for one trajectory (60 turns; 208 trajectories; 2,496 questions). We evaluate \bcpm{} and \ama{} answers against references using gpt-5.6-sol~\citep{openai2026gpt56sol}. For \bcpm{}, we follow the BrowseComp grading rubric~\citep{wei2025browsecomp}.

\paragraph{Samples and re-evaluation.}
Evaluations are conducted every ten training steps at a sampling temperature of 0.7.
ReAct, masking and MemGPT have two samples per checkpoint.
Our first evaluation harness discarded a compaction summary whenever the model answered the compaction request with a tool call and wrote the summary in its reasoning. Training was unaffected because it used a different serving path. We fixed the evaluation harness and re-evaluated \supo and \method at every checkpoint. For \method on all benchmarks and \supo on \bcpm and \ama file, we report only these re-evaluation results.
\supo's compactions on \cb and \ama stream offer no tools, so its two samples per checkpoint there are unaffected.
Masking, MemGPT and ReAct (which never compacts) were evaluated with the fixed harness.
For the reported RL-trained models, we drew one more sample for every strategy except ReAct, whose selected checkpoint was not retained, and two more for \supo and \method on the re-evaluated benchmarks, so each item is averaged over three samples (ReAct: two).
Costs after RL come from one of these samples, run for every strategy with the same settings (eight GPUs, up to 256 concurrent episodes); MemGPT's third sample was split across several servers, so its costs are the mean of its two regular evaluations, which used the same settings.
Confidence intervals come from 10{,}000 paired bootstrap resamples of the evaluation items (questions, episodes, streams or trajectories). For \ama question types we resample trajectories.

\section{Additional Results}
\label{app:more}

\paragraph{Single checkpoints.}
Single checkpoints fluctuate by several points (Figure~\ref{fig:curves}): \supo on \bcpm drops from 42.9 at the preceding evaluation to 38.1 for the reported RL-trained model, and \method on \cb Log Triage drops from 97.3 to 88.1 between two earlier evaluations.

\paragraph{\ama stream by length and question type.}
Table~\ref{tab:ama} breaks \ama stream down after RL.
\method is within 2.3 points of \supo on both length buckets and all four question types, with no statistically significant differences. Its largest gain is on trajectories longer than the window (+2.3 \ci{$-$0.2}{4.9}).

\begin{table}[ht]
\centering
\small
\setlength{\tabcolsep}{4pt}
\resizebox{\linewidth}{!}{%
\begin{tabular}{@{}lcccc@{}}
\toprule
Subset & \supo & Mask. & MemGPT & \method \\
\midrule
$\le$ 32K tokens (124) & 48.2 & \textbf{48.3} & 47.9 & \textbf{48.3} \\
32K--128K tokens (65) & 39.4 & 38.3 & 40.6 & \textbf{41.7} \\
\midrule
A: recall & 48.8 & 48.3 & 49.2 & \textbf{50.0} \\
B: causal inference & 42.2 & 44.1 & \textbf{45.1} & 43.1 \\
C: state updating & 51.7 & 49.4 & 49.8 & \textbf{53.6} \\
D: state abstraction & 31.9 & \textbf{32.0} & 31.1 & 30.4 \\
\bottomrule
\end{tabular}}
\caption{\textbf{AMA-Bench Stream Accuracy by Length and Question Type.} We break down accuracy (\%) after RL training by trajectory length and question type. Length groups are defined by whether trajectories fit within the 32K-token context window, with trajectory counts shown in parentheses. Question types cover recall, causal inference, state updating, and state abstraction. Bold text marks the best value in each row.}
\label{tab:ama}
\end{table}

\begin{table}[H]
\centering
\small
\setlength{\tabcolsep}{3pt}
\resizebox{\linewidth}{!}{%
\begin{tabular}{@{}lcccc@{}}
\toprule
Benchmark & \supo & Masking & MemGPT & \method \\
\midrule
\bcpm      & 1.15M / 84  & 949K / 51  & 1.97M / 101 & 1.22M / 88 \\
\ama file  & 590K / 41   & 490K / 34  & 674K / 38   & 560K / 38 \\
\ama stream& 377K / 24   & 404K / 23  & 1.30M / 67  & 869K / 56 \\
\cb        & 2.11M / 114 & 2.41M / 112 & 8.60M / 391 & 2.51M / 154 \\
\bottomrule
\end{tabular}}
\caption{\textbf{Token Usage and LLM Calls After RL.} We report the average token usage and number of LLM calls per episode for each strategy and benchmark. Each entry shows processed tokens / LLM calls, with tokens including prompt and generated tokens summed over all calls. ContextBench costs are measured per stream. ReAct is omitted because its selected checkpoint was not retained.}
\label{tab:cost}
\end{table}

\paragraph{Cost.}
Table~\ref{tab:cost} lists the costs behind Figures~\ref{fig:teaser} and~\ref{fig:cost}, and Figure~\ref{fig:costtime} plots accuracy against wall-clock time per episode instead of tokens.
\method also lies on the accuracy--runtime Pareto frontier
for every task. On \cb, it is faster than \supo, taking
38 versus 61 minutes per KV Store stream and 56 versus
84 minutes per Log Triage stream, as it generates shorter
summaries. We could not measure ReAct's cost because its
selected checkpoint was not retained. ReAct stops when
its context window is full.